# LDOP: Local Directional Order Pattern for Robust Face Retrieval

Shiv Ram Dubey and Snehasis Mukherjee

Computer Vision Group, Indian Institute of Information Technology, Sri City, Chittoor, A.P.-517646, India

srdubey@iiits.in, snehasis.mukherjee@iiits.in

*Abstract—* **The local descriptors have gained wide range of attention due to their enhanced discriminative abilities. It has been proved that the consideration of multi-scale local neighborhood improves the performance of the descriptor, though at the cost of increased dimension. This paper proposes a novel method to construct a local descriptor using multi-scale neighborhood by finding the local directional order among the intensity values at different scales in a particular direction. Local directional order is the multi-radius relationship factor in a particular direction. The proposed local directional order pattern (LDOP) for a particular pixel is computed by finding the relationship between the center pixel and local directional order indexes. It is required to transform the center value into the range of neighboring orders. Finally, the histogram of LDOP is computed over whole image to construct the descriptor. In contrast to the state-of-the-art descriptors, the dimension of the proposed descriptor does not depend upon the number of neighbors involved to compute the order; it only depends upon the number of directions. The introduced descriptor is evaluated over the image retrieval framework and compared with the state-of-the-art descriptors over challenging face databases such as PaSC, LFW, PubFig, FERET, AR, AT&T, and ExtendedYale. The experimental results confirm the superiority and robustness of the LDOP descriptor.**

*Index Terms—* **Local descriptors, Face image, Local ordering, Intensity order, Image retrieval, Robustness, LBP.**

## 1. INTRODUCTION

Facial image analysis such as face recognition, face retrieval and facial expression recognition is being widely studied by the researchers during the last few decades. Recently, face recognition techniques in the unconstrained environment is gaining much interest of the researchers, where either the images are collected from the internet such as "Labeled Faces in the Wild" [1] and "Public Figures" [2] or, the images are taken from the surveillance cameras and mobile devices [3]. The major problem with face image analysis is the sensitivity to the image capturing environments such as pose variations, scale change, illumination difference, blurring effect, viewpoint change, and noise [4], [5]. To deal with the face images with these geometric and photometric challenges, it is required to develop robust methods capable of handling adverse imaging conditions. The advancement and recent trends in the effective solution for face image analysis have been extensively reviewed by several researchers from time to time [6], [7], [8].

Any face recognition approach mainly has two components: face description and face matching using the multi-class classifiers such as nearest neighbor (NN) classifier and sparse representation classifier (SRC) [9]. The image feature description has been proved to be the backbone in most of the computer vision problems. An efficient feature descriptor is expected to have the following three properties: a) distinctive, b) robust, and c) low dimensional. In this paper, a local directional order pattern (LDOP) based face representation is proposed for face retrieval as illustrated in Fig. 1. First, the intensity order among neighbors at different radius in a particular direction is used to get the local directional orders. Then, the original image is transformed into the range of local directional orders. After that, the proposed descriptor is computed by encoding the relationship between transformed image and local directional orders. At last, the most similar faces are retrieved from the database on the basis of the distances between the descriptors of the query image and database images.

The rest of paper is organized in following manner: Section 2 discusses the related works; Section 3 introduces the construction process of proposed LDOP descriptor as well as multi resolution LDOP descriptor; Section 4 presents the face retrieval framework using multi resolution LDOP descriptor; The experimental analysis over benchmark face databases are carried out in Section 5; and finally, Section 6 concludes the paper with motivating remarks about LDOP descriptor.

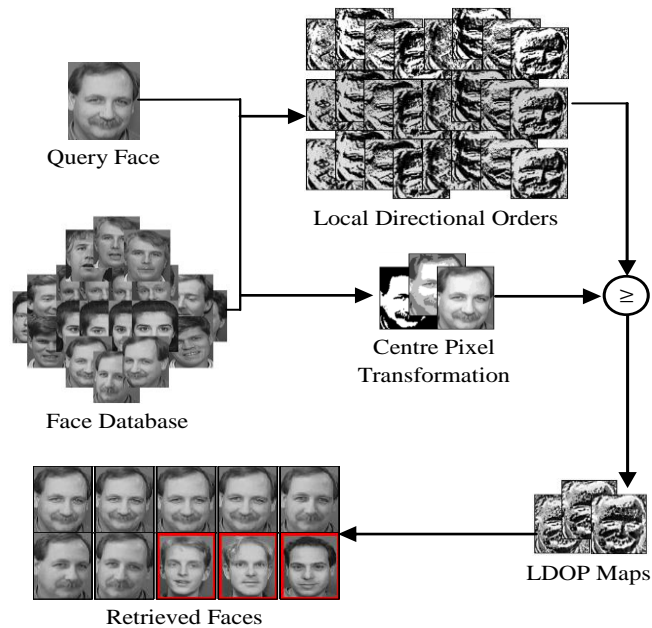

Fig.1. Face retrieval framework using multi-resolution LDOP descriptor. False positives are enclosed in red rectangles. Images are taken from AT&T database [52].

## 2. RELATED WORKS

Face description is required to make the face recognition approaches robust to intra-class variations and discriminative to inter-class similarities. The face image-descriptor based methods [10], [11], [12], [13], and deep learning based methods [14], [15], are the two major approaches widely adapted in the research community, for face retrieval. The advantages of former methods are data independence, ease of use, no complex computation facility needed and robustness to real-situation variations like rotation, scale, expression and illumination differences.

The image-descriptor based methods can be further classified into two categories, i.e., hand-crafted descriptor and learning based descriptor. The designing of hand-crafted image descriptors are the mostly followed research area for the face representation. The local binary pattern (LBP) is proved as a very efficient and simple feature descriptor to capture the micro information of the image [16], [17]. The LBP based approaches have shown promising performance for different computer vision applications such as texture classification and retrieval [18], [19], [20], [21]. Ahonen et al. investigated the suitability of LBP for the face recognition task [10]. They computed the LBP descriptor over several blocks and concatenated to form a single descriptor. This approach outperformed the classical methods such as PCA, Bayesian Classifier and Elastic Bunch Graph Matching. Recently, Huang et al. surveyed the LBP based approaches for face recognition [22] and Yang and Chen [23] have presented a comparative study over LBP based face recognition techniques. Several local descriptors have been investigated for face recognition inspired by the success of LBP [13], [24], [25], [26], [27], [28], [29], [30], [31], [32], [33], [34]. Zhang et al. modeled a Local Gabor Binary Pattern Histogram Sequence (LGBPHS) by concatenating the histograms of all the local regions of all the local Gabor magnitude binary pattern maps for a face image [24]. The LGBPHS consists of histograms at different scale and orientation which make it more effective than LBP. The local ternary pattern (LTP) is proposed for illumination robust face recognition by considering ternary value against binary value of LBP and shown promising performance [25]. The LTP descriptor is more discriminant and less sensitive to noise in uniform regions. Zhang et al. did a revolutionary work over LBP and considered LBP over high-order derivatives of the image to introduce local derivative pattern (LDP) for face representation [26]. The pattern at high-order local derivative captures more detailed information as compared to the first order local pattern. Weber local descriptor (WLD) is proposed for faces and texture recognition with inspiration from the Weber's theory [27]. Local Gabor XOR pattern (LGXP) is introduced by utilizing the Gabor magnitude and phase information in the LBP framework [28]. The LBP feature on orientated edge magnitudes is computed to form the Patterns of Oriented Edge Magnitudes (POEM) descriptor [29]. Multiscale local phase quantization (MLPQ) is proposed for blur-robust face recognition by encoding the blur-invariant information in face images [30]. Local directional number pattern (LDN) used the prominent direction computed from Kirsch masks to get the pattern [31]. The LDN does not utilize the wider neighborhood, whereas the proposed approach does. The LDN requires the computation of edge response, whereas our approach requires the computation of directional order. Recently, the concept of LDP descriptor is extended by Fan et al. into local vector pattern (LVP) for face representation [32]. They encoded the relationship of referenced pixel with neighbouring pixels in different directions at different distances over high-order face image. The proposed Local Directional Order Pattern (LDOP) is different from the Local Vector Pattern (LVP). The proposed LDOP descriptor encodes the directional ordering pattern, whereas the LVP descriptor encodes the relationship between the differences of intensity at different directions. Unlike the LDOP descriptor, the LVP descriptor works on high-order image space. The only similarity between both is that both incorporate the directional information. The unique novelty of LDOP is that it uses the intensity order along different directions to form the descriptor. Basically, it encodes the relationship among the directional order pattern. The semi-structure local binary pattern (SLBP) derived the LBP over locally aggregated image (i.e. the center is replaced by sum of intensities from its 3×3 window) for facial analysis [33]. Other notable descriptors for face recognition include local directional gradient pattern (LDGP) [34] and dual cross patterns (DCP) [13]. The relationships between higher order derivatives in four directions are utilized to encode the LDGP [34]. DCP works by computing the two cross patterns with different set of neighbors selected from two different radiuses [13]. This work uses the first derivative of Gaussian operator (FDG) to convert the face image into multi-directional gradient image which in terns increases the dimension of the final descriptor [13]. Arandjelovi has proposed the gradient edge map features for face recognition, where he directly compared the feature images instead of forming the descriptor [35]. This method can be used only for the frontal images.

A harvesting visual concept is used in [56] for image search with complex queries. Researchers have also explored the Wavelet-based feature descriptors for different applications such as image retrieval and watermarking [57], [58], [59]. In recent years, several learning-based descriptors using supervised and unsupervised techniques have evolved. Some well-known descriptors in this category are Learning-based (LE) Descriptor [11], Learned Background Descriptor [12], AdaBoost Multi-scale Block LBP [36], Local Quantized Patterns (LQP) [37], Discriminant Face Descriptor (DFD) [38], and Compact Binary Face Descriptor (CBFD) [39]. The learning-based descriptors are having the wider diversity in terms of the larger sampling size as compared to the hand-crafted descriptors.

There are two kinds of order based descriptors, one using LBP framework in high-order derivative space and another using intensity order among neighboring values for local region matching. Some examples of former approach are LDP [26] and LDGP [34]. Recently, the intensity order based descriptors have been introduced for feature description and applied mainly to the local region matching [40], [41], [42], [43]. The Local Intensity Order Pattern (LIOP) is widely known due its simplicity and robustness for region matching [41]. Using intensity order as compared to raw intensity values provides the inherent robustness to the descriptor over uniform illumination changes. The LBP descriptor [10] has been successful in several computer vision applications. It has been also observed that the performance of LBP can be boosted further by considering it at the higher scales at the cost of high dimensionality. The LBP is suitable for capturing the micro information, whereas it fails to capture the macro information from the image due to the limited local neighbourhood. Some LBP variants consider the multiple scales but do not utilize the relationship among neighbours at different scales which leads to high dimensionality.

In this work, we utilize the properties of intensity order over directional neighbors to solve the following two problems: to encode the wider neighborhood to discriminate the inter-class similarity and to have the robustness for the intra-class variations. Next we illustrate the process of constructing the proposed LDOP descriptor.

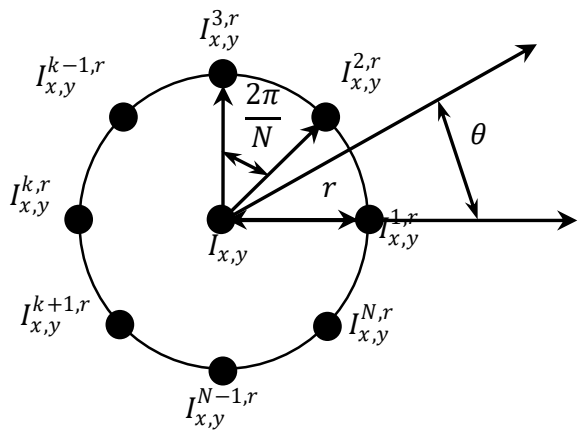

Fig.2. The $N$ equally spaced local neighbors $(x_r^k, y_r^k)|_{\forall\ k\in[1,N]}$ at a radius of $r$ from a center pixel $(x, y)$.

## 3. LOCAL DIRECTIONAL ORDER PATTERN

We describe the proposed LDOP descriptor first, followed by the multi-resolution LDOP descriptor.

### A. Proposed LDOP Descriptor

In this sub-section, the construction process of local directional order pattern is described. First, we discuss the local directional neighborhood extraction, then local directional order computation, then center pixel transformation, and finally the construction of local directional order pattern.

*Local Directional Neighbors*

Let $I$ is an image with dimension $d_x \times d_y$ and having a bit-depth of $B$ bits. Let $I_{x,y}$ represents the intensity value at co-ordinate $(x,y)$, where $1 \leq x \leq d_x$, $1 \leq y \leq d_y$ and $0 \leq I_{x,y} \leq (2^B - 1)$. Let $I_{x,y}^{k,r}$ is the intensity value of $k^{th}$ neighbor of the center pixel $(x,y)$ when total $N$ neighbors are considered equally spaced at a radius of $r$ from the center as shown in Fig. 2. Let $(x_r^k, y_r^k)$ is the co-ordinates of $I_{x,y}^{k,r}$, then we can also say that $I_{x,y}^{k,r} = I_{x_r^k, y_r^k}$. The co-ordinates of $k^{th}$ neighbor at radius $r$ (i.e., $x_r^k$ and $y_r^k$) is given as follows,

$$x_r^k = x - r(\sin(\theta_k)), \tag{1}$$

$$y_r^k = y + r(\cos(\theta_k)), \tag{2}$$

where $\theta_k|_{1 \leq k \leq N}$ is the angular displacement of $k^{th}$ neighbor from $1^{st}$ neighbor in counter-clockwise direction, w.r.t. center pixel and computed as follows,

$$\theta_k = (k-1)\frac{2\pi}{N}. \tag{3}$$

It is obvious that $\theta_k$ represents the $k^{th}$ direction w.r.t. the center. We refer the intensity value of $r^{th}$ neighbor (i.e., neighbor at radius $r$ from the center) in $k^{th}$ direction by $I_{x,y}^{k,r}$. We define $R$ directional neighbors of center $(x,y)$ in $k^{th}$ direction by a vector $P_{x,y,R}^k$. So, $P_{x,y,R}^k$ will consist of the following neighboring values,

$$P_{x,y,R}^k = (I_{x,y}^{k,1}, I_{x,y}^{k,2}, \dots, I_{x,y}^{k,r}, \dots, I_{x,y}^{k,R}). \tag{4}$$

*Local Directional Orders*

In multi-scale descriptors like LVP [32], the relationship among neighbors at different radius is not encoded. In this work, we encode the relationship among the local neighbors at different radius (i.e., distance from the center) in a particular direction to encode the directional relationship among neighbors.

Let $S_{x,y,R}^k$ is the sorted version of $P_{x,y,R}^k$ in the increasing order. We represent $S_{x,y,R}^k$ as follows,

$$S_{x,y,R}^k = (s_{x,y}^{k,1}, s_{x,y}^{k,2}, \dots, s_{x,y}^{k,r}, \dots, s_{x,y}^{k,R}). \tag{5}$$

The order in the $k^{th}$ direction is represented by $O_{x,y,R}^k$ and defined as follows,

$$O_{x,y,R}^k = (o_{x,y}^{k,1}, o_{x,y}^{k,2}, \dots, o_{x,y}^{k,r}, \dots, o_{x,y}^{k,R}), \tag{6}$$

such that

$$I_{x,y}^{k,r} = s_{x,y}^{k,o_{x,y}^{k,r}}, \tag{7}$$

where

$$o_{x,y}^{k,r} \in [1, R]|_{\forall r \in [1,R]},$$

and

$$o_{x,y}^{k,i} \neq o_{x,y}^{k,j}|_{i \neq j}.$$

Now, we are having $R$ distinct values in $k^{th}$ directional order $O_{x,y,R}^k$ for which we have to compute an index value $\omega_{x,y,R}^k$ to represent the order using a single value.

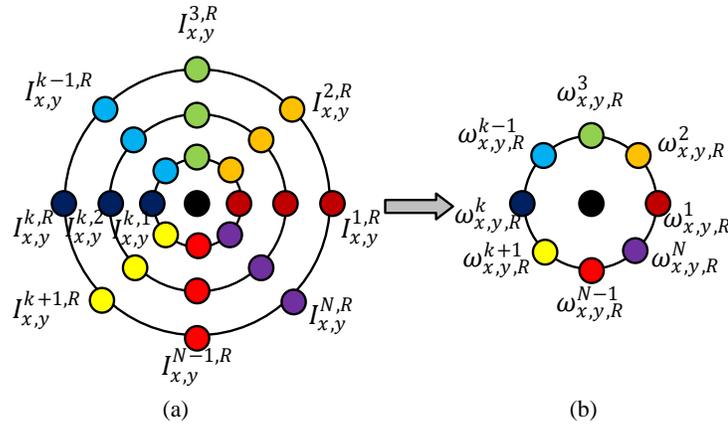

(a)     (b)

Fig.3. The proposed idea of encoding the relationship among local neighbors at different radius using the order: (a) the local neighbors $I_{x,y}^{k,r}|_{\forall\, k\in[1,N]\,\&\, r\in[1,R]}$ of a center pixel $(x,y)$ and (b) the $R$ neighbors in $k^{th}$ direction (i.e. $I_{x,y}^{k,r}|_{\forall\, r\in[1,R]}$) converges to a single value $\omega_{x,y,R}^{k}$ which actually represents the order among the corresponding neighbors in (a).

Let $\varphi$ represents all the permutations of the $R$ distinct values between 1 and $R$ in lexicographical fashion. The total possible number of permutations with $R$ values are $R!$. The $\varphi_i|_{i\in[1,R!]}$ (i.e., the $i^{th}$ permutation) is having the following values,

$$\varphi_i = (\varphi_i^1, \varphi_i^2, \ldots, \varphi_i^r, \ldots, \varphi_i^R). \tag{8}$$

The order index $\omega_{x,y,R}^{k}$ in $k^{th}$ direction with $R$ radius for pixel $(x,y)$ is defined as follows,

$$\omega_{x,y,R}^{k} = j, \tag{9}$$

such that

$$\varphi_j^r = o_{x,y}^{k,r}|_{\forall r\in[1,R]}.$$

This idea is illustrated in Fig. 3, where by encoding the relationship exist among the $R$ neighbors in $k^{th}$ direction (i.e., $I_{x,y}^{k,r}|_{\forall\, r\in[1,R]}$) in terms of its order, it converges to a single value $\omega_{x,y,R}^{k}$ which represents the order index for the neighboring values in $k^{th}$ direction. The LDOP descriptor inherently become uniform illumination robust, inherently as the directional order among the directional neighbors is uniform illumination invariant. Fig. 4(f-h) display the local directional order map (i.e., $\omega_R^k$) for $R=1,2,\,\&\,3$ respectively, computed over an input example face image depicted in Fig. 4(a). The eight columns in Fig. 4(f-h) correspond to local directional order map for 8 directions staring from $0^o$ with an intervals of $45^o$ (i.e., $\omega_R^k|_{k=1,2,\ldots,8}$) respectively. The range of values in images of Fig. 4(f-h) are $1-2, 1-6,\,\&\,1-24$ respectively, because the maximum possible order index value is $R!$. From the images of Fig. 4(f-h), it is clear that local directional orders capture the useful directional information for designing the LDOP descriptor.

*Center Pixel Transformation*

As of now, we have computed $N$ local directional order index values (i.e., $\omega_{x,y,R}^{k}|_{k=1,2,3\ldots,N}$) from the local neighborhoods of the center pixel. In other words, the directional relationship among the local neighbors is coded into $\omega_{x,y,R}^{k}$. Our next goal is to make a low dimensional descriptor as well as to utilize the relationship of center with its neighbors. At this point, it is difficult to compare the center $I_{x,y}$ with local directional orders $\omega_{x,y,R}^{k}$ due to the range mismatch. It can be noted that the range of $I_{x,y}$ is $[0, 2^B - 1]$ for $B$ bit-depth, whereas the range of $\omega_{x,y,R}^{k}$ is $[1, R!]$. To resolve this issue, a center pixel transformation scheme is developed in this paper to transform the range of $I_{x,y}$ into the range of $\omega_{x,y,R}^{k}$. Let $T_{x,y,R}$ is the transformed version of $I_{x,y}$ when $R$ directional neighbors are used to generate $\omega_{x,y,R}^{k}$. The $T_{x,y,R}$ is computed as follows,

$$T_{x,y,R} = \frac{I_{x,y} \times (R! - 1)}{2^B - 1} + 1. \tag{10}$$

Note that, this center pixel transformation scheme increases the tolerance capability of the descriptor towards noise for smaller values of $R$ such as 2, 3 and 4.

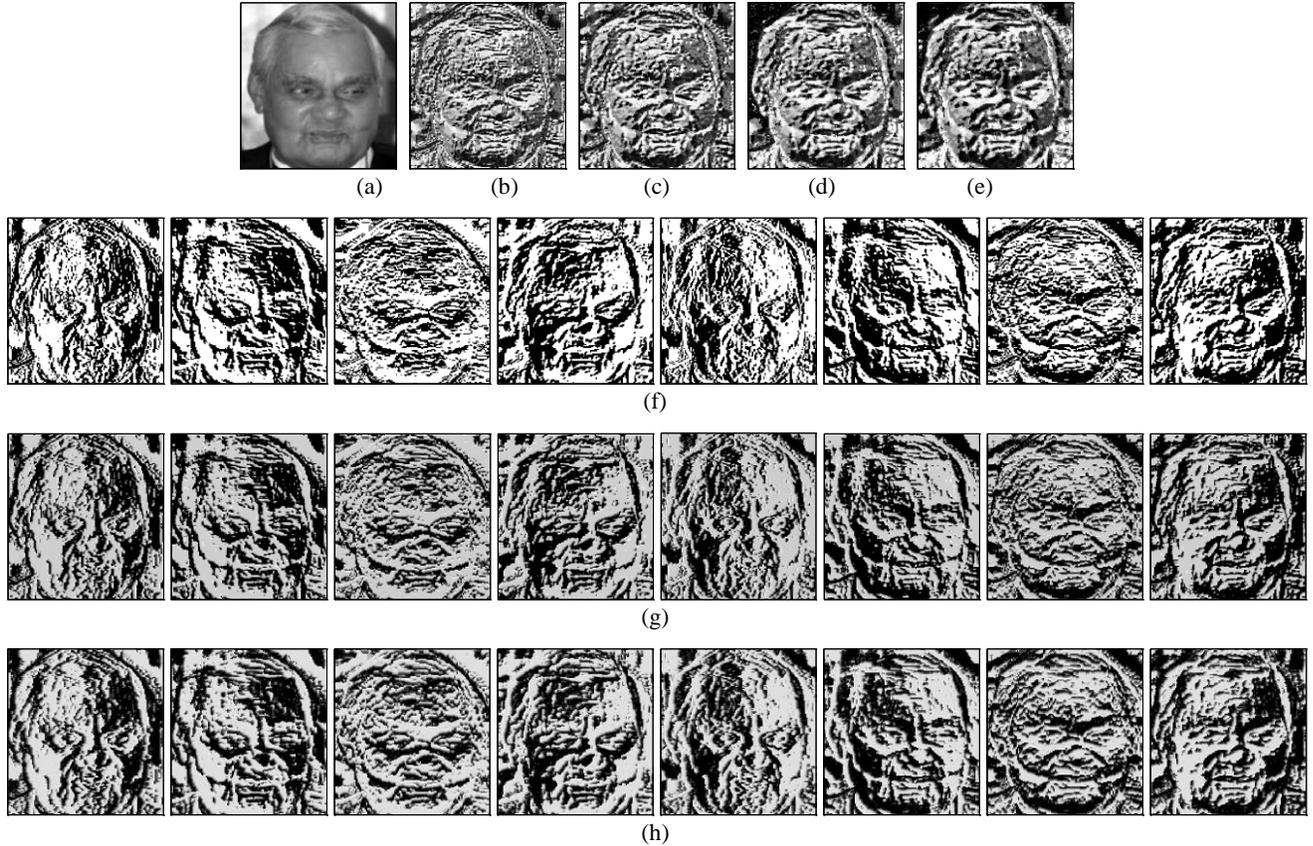

Fig.4. The illustration of the proposed concept using an example image from LFW database [47] with 8 local neighbors: (a) Input Image, $I$, (b) LBP map, (c) LDOP map for $R = 2$, (d) LDOP map for $R = 3$, (e) LDOP map for $R = 4$, (f) local directional order maps for $R = 2$ in 8 directions 0, 45, 90, 135, 180, 225, 270, and 315 degrees respectively, (g) local directional order maps for $R = 3$ in same 8 directions respectively, and (h) local directional order maps for $R = 4$ in the same 8 directions respectively.

*LDOP Descriptor*

The $\omega_{x,y,R}^k|_{k=1,2,3…,N}$ and $T_{x,y,R}$ are used to encode the relationship of transformed center with its directional neighbors converted into orders. The local directional order pattern map ($LDOPm$) for center $I_{x,y}$, is defined as follows,

$$LDOPm_{x,y,R} = \sum_{k=1}^{N} w_k \times \delta_{x,y,R}^k, \quad (11)$$

where $R$ is the radius for neighbors and $w_k$ is a weighting factor for the $k^{th}$ direction and determined as follows,

$$w_k = (2)^{(k-1)}, \quad (12)$$

and $\delta_{x,y,R}^k$ is the binary relationship factor between center and its $k^{th}$ directional neighbors and computed as follows,

$$\delta_{x,y,R}^k = \begin{cases} 1, & if\ \omega_{x,y,R}^k \geq T_{x,y,R} \\ 0, & Otherwise \end{cases}. \quad (13)$$

The LDOP maps over the input image considered in Fig. 4(a) for $R = 2, 3, \& 4$ are displayed in Fig. 4(c-e). The LBP map over the same image is displayed in Fig. 4(b). It can be observed from the Fig. 4(b-e) that the LBP is encoding too much useless information, whereas, LDOP is avoiding useless information. Moreover, it can be noted that LDOP is capturing more detailed information for lower value of $R$ using narrow neighborhood and capturing more corus information for higher value of R using wider neighborhood.

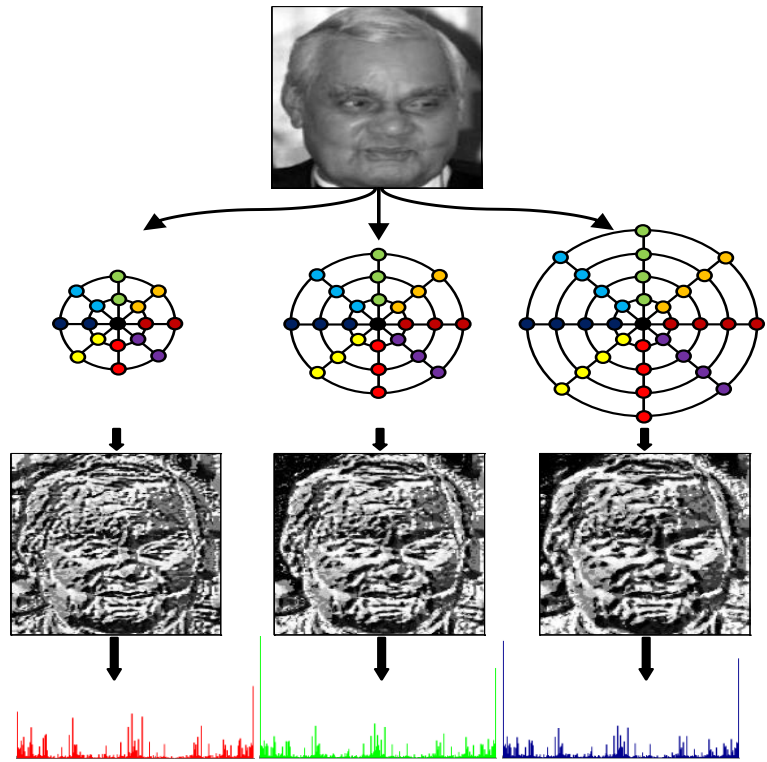

Fig.5. Procedure to construct multi-resolution $LDOP$ descriptor (i.e. $LDOP_{2,4}$) with $R = 2, 3, \& 4$.

The final $LDOP$ descriptor is the histogram computed over $LDOP$ map (i.e. $LDOPm$) as follows,

$$LDOP_R(\zeta) = \frac{1}{d_x^R \times d_y^R} \sum_{x=R+1}^{d_x-R} \sum_{y=R+1}^{d_y-R} \xi(LDOPm_{x,y,R}, \zeta), \quad (14)$$

$\forall \zeta \in [0, 2^N - 1]$, where, $LDOP_R$ represents the $LDOP$ descriptor using radius $R$ neighborhood, $d_x^R = d_x - 2R$, $d_y^R = d_y - 2R$, and $\xi(\alpha, \beta)$ is a function defined as follows,

$$\xi(\alpha, \beta) = \begin{cases} 1, & if \ \alpha = \beta \\ 0, & Otherwise \end{cases}. \quad (15)$$

The dimension of $LDOP$ descriptor depends upon the number of directions considered and given by $2^N$. The eight directions are considered in the experiments of this paper, thus the dimension of $LDOP$ is 256.

**B. Multi Resolution LDOP**

The multi resolution $LDOP$ accumulates the $LDOP$ descriptors at different scales/radius. It actually exhibits the progressive effect of increasing the area of local neighborhood over the $LDOP$ descriptor. We define the multi resolution $LDOP_{R1,R2}|_{R1>1, R2>1 \ \& \ R1 \leq R2}$ as follows,

$$LDOP_{R1,R2} = [LDOP_{R1}, LDOP_{R1+1}, \dots, LDOP_{R2}] \quad (16)$$

The dimension of multi resolution $LDOP$ is given by $(R2 - R1 + 1) \times 2^N$. The construction of multi resolution $LDOP$ with 2, 3 and 4 radius of neighborhoods is illustrated in Fig. 5 using the same example image considered in Fig. 4(a). The $LDOP$ maps at different radius (i.e. $LDOPm_2$, $LDOPm_3$, and $LDOPm_4$) are derived first and then the feature vector using histograms over each map is concatenated to form the final $LDOP$ multi resolution descriptor, which is used for face retrieval.

## 4. FACE RETRIEVAL USING LDOP

The framework for face retrieval using the proposed multi resolution LDOP is depicted in Fig. 1. For query image as well as database images, LDOP based descriptors are computed. The LDOP descriptor of query face is compared with all the faces of the database by finding the distances between them. Based upon the lower distances, top $\gamma$ faces are retrieved from the database. In this paper, Chi-square distance is used in most of the experiments, whereas the performance of proposed descriptor with other distances like Euclidean distance, Cosine distance, $L_1$ distance, and $D_1$ distance is also investigated [20], [44].

### A. Evaluation Criteria

In face retrieval, the ultimate goal is to retrieve the most similar faces against a query face from a face database. For evaluation purpose, we converted each face image of the database into the query image and retrieved top $\gamma$ faces from the database. Note that, the query face image is also present in the database, thus, the first retrieved face is the query face itself. The *Precision* and *Recall* metrics are used to evaluate the performance of different descriptors for face retrieval. The average retrieval precision (ARP) and average retrieval rate (ARR) for a given database are defined as follows,

$$ARP = \frac{\sum_{i=1}^{\mathbb{C}} AP(i)}{\mathbb{C}}, \quad (17)$$

$$ARR = \frac{\sum_{i=1}^{\mathbb{C}} AR(i)}{\mathbb{C}}, \quad (18)$$

where $\mathbb{C}$ represent the total number of classes in the database, $AP(i)$ and $AR(i)$ represent the average precision and average recall respectively over the images of the $i^{th}$ class of the database and given as follows,

$$AP(i)|_{\forall\ i\in[1,\mathbb{C}]} = \frac{\sum_{j=1}^{\mathbb{C}_i} Pr(j)}{\mathbb{C}_i}, \quad (19)$$

$$AR(i)|_{\forall\ i\in[1,\mathbb{C}]} = \frac{\sum_{j=1}^{\mathbb{C}_i} Re(j)}{\mathbb{C}_i}, \quad (20)$$

where $\mathbb{C}_i$ represent the number of face images in the $i^{th}$ class of the database, $Pr$ and $Re$ represent the precision and recall respectively for a query face and defined as follows,

$$Pr(k)|_{\forall\ k\in[1,\mathbb{C}_j]} = \frac{\#\ of\ retrieved\ similar\ images}{\gamma}, \quad (21)$$

$$Re(k)|_{\forall\ k\in[1,\mathbb{C}_j]} = \frac{\#\ of\ retrieved\ similar\ images}{\#\ of\ similar\ images\ in\ database}. \quad (22)$$

where $\gamma$ represent the number of retrieved faces when $k^{th}$ image of $i^{th}$ class is considered as the query face.

We also computed the F-Score from ARP and ARR in terms of the number of top matching retrieved images. The average normalized modified retrieval rank (ANMRR) metric is also computed and represented in percentage in this paper [45]. The higher value of ARP, ARR and F-Score means the better retrieval performance and vice-versa, whereas the lower value of ANMRR represents the better retrieval performance and vice-versa.

### B. Face Databases

We have used seven face databases namely PaSC [46], LFW [1], [47], PubFig [2], FERET [48], [49], AR [50], [51], AT&T [52] and ExtendedYale [53], [54], for the face retrieval. After localization, all the face images are down-sampled in 64×64 dimension. The PaSC still images face database is very challenging due to many variations present such as pose, illumination and blur [46]. Total 9376 images from 293 subjects with 32 images per subject are present in the PaSC database. We used Viola Jones object detection method [55] for face localization over PaSC images. Finally, we have 8718 faces in this database where faces are successfully detected using Viola Jones detector.

The face retrieval in unconstrained environment is a fundamental problem. LFW and PubFig databases contain the images collected from the internet. These images are captures in completely unconstrained environments with non-cooperative subjects. Thus, the variations like pose, lighting, expression, scene, camera, etc. are present. The gray-scaled version of LFW cropped database [47] is used in this paper. In image retrieval framework, it is required to retrieve more than one (typically 5, 10, etc.) best matching images. So, in our database, the sufficient number of images should be present in each category. Thus, we have considered only those subjects that are having at least 20 images. Finally, in this database, there are 3023 face images from 62 individuals. The PubFig (i.e. PublicFigure) database is having 60 individuals with 6472 number of total images in the database [2]. We have downloaded the images from the internet directly following the urls given in this database and removed the dead urls.

"Portions of the research in this paper use the FERET database of facial images collected under the FERET program, sponsored by the DOD Counterdrug Technology Development Program Office" [48], [49]. Due to the severe variations in the expression and pose (13 different poses), Color-FERET database is adopted as a challenging database. We considered only those subjects that are having at least 20 images, converted the color images into gray-scale images. Finally, FERET database is having 141 subjects with 4053 total number of images. The AR face databases exhibit different facial expressions, illumination conditions and occlusions [50]. We have used the cropped version of this database [51] and converted the color image into gray-scaled image. The AR database consists of a total 2600 images from 100 individuals with 26 images per individual. The AT&T face database (formerly "The ORL Database of Faces") is having 10 images per subject from 40 different subjects [52]. There are variations in the lighting conditions and facial expressions for some subjects. A dark homogeneous background is used to capture the images in AT&T database with the subjects in an upright, frontal position. The ExtendedYale database is having gray-scaled cropped and aligned faces [53], [54]. This database is created in the controlled lighting environment having severe and non-uniform illumination changes. Total 38 subjects are present in the database. Each subject is having 64 images with 2432 total images in this database.

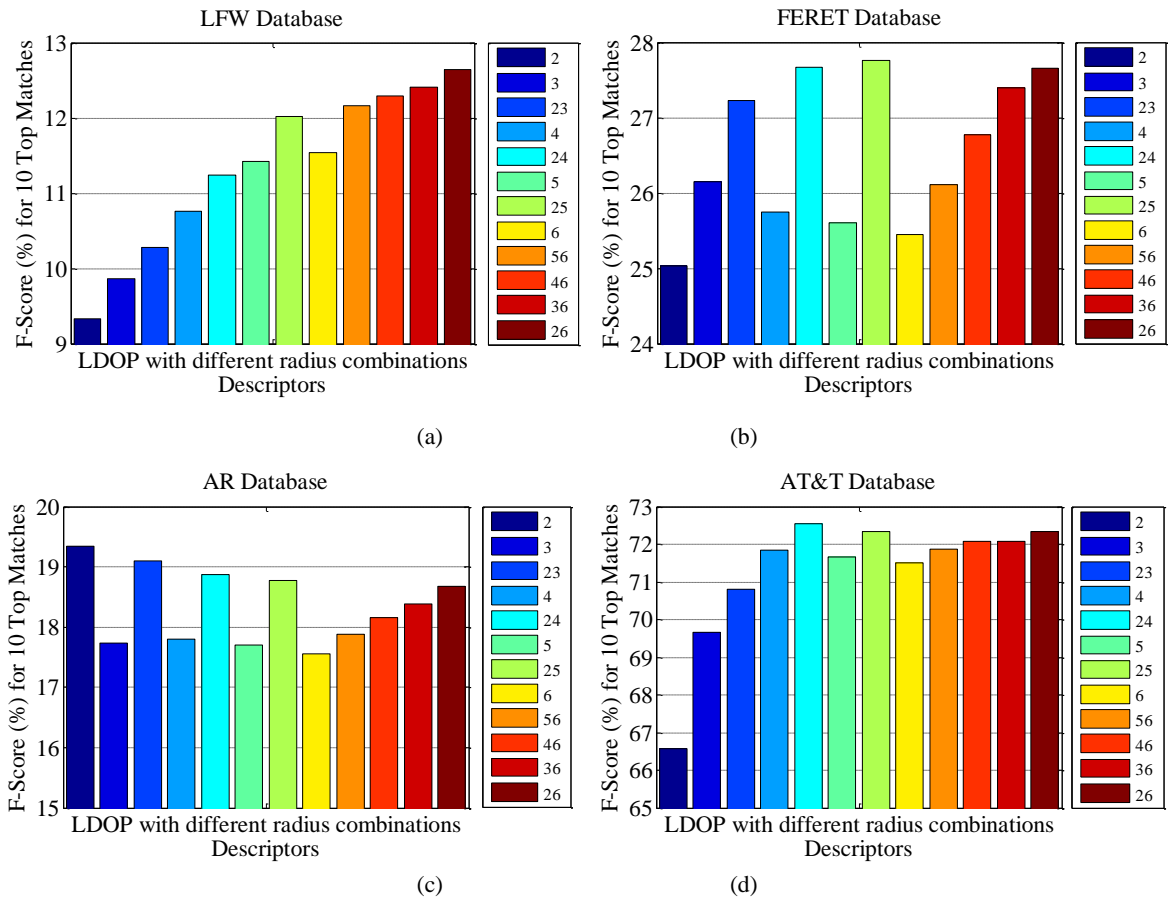

Fig.6. The F-Score (%) when 10 images are retrieved ($\gamma = 10$) over (a) LFW, (b) FERET, (C) AR, and (d) AT&T databases using LDOP descriptor with different radius and multi resolution. In the legends, the radius or range of radius (for multi resolution) is mentioned; the single digit represents a radius, whereas double digit represents the range of radius.

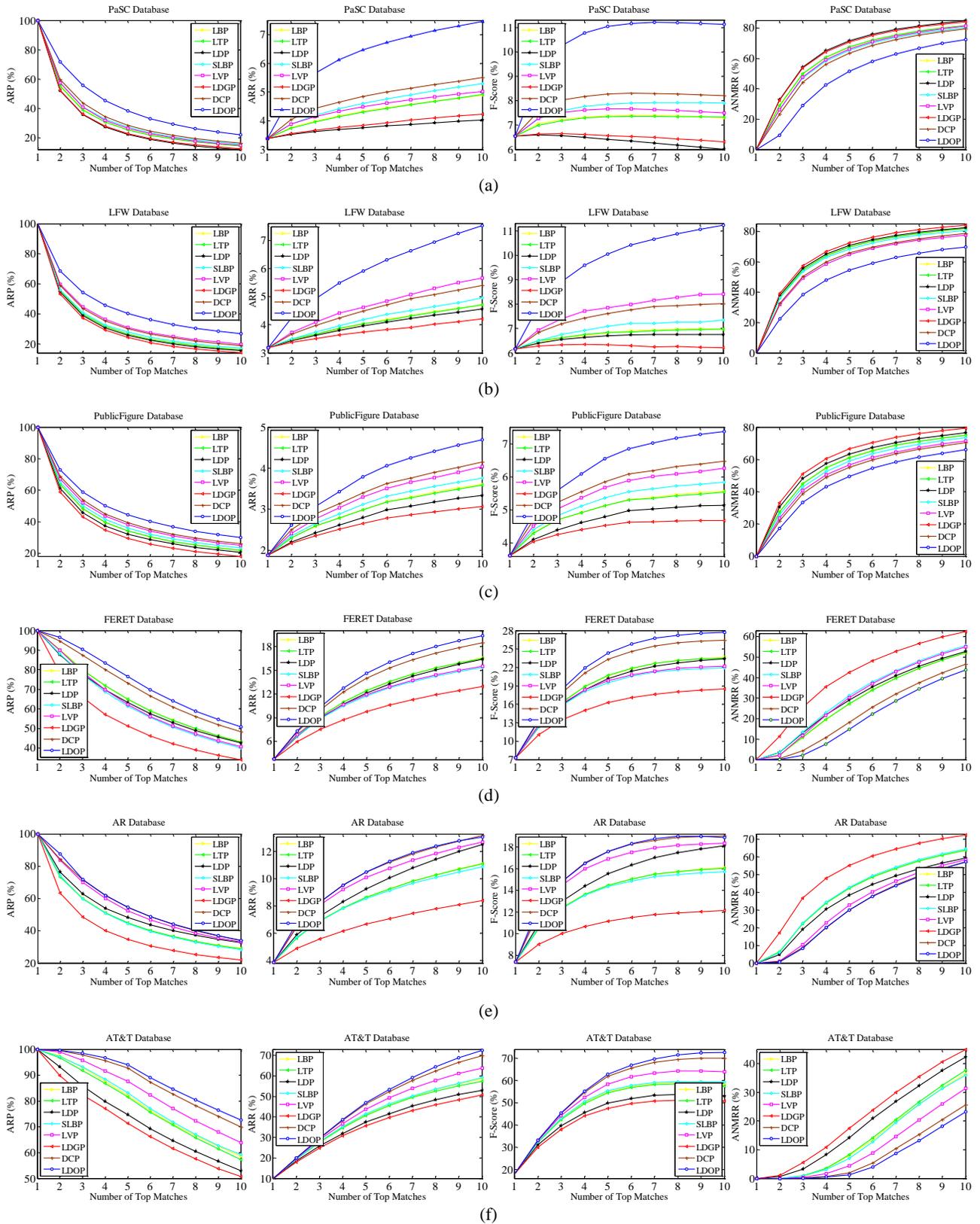

Fig.7. The experimental retrieval results in terms of the ARP (%), ARR (%), F-Score (%), and ANMRR (%) in the $1^{st}$, $2^{nd}$, $3^{rd}$, and $4^{th}$ columns over (a) PaSC, (b) LFW (c) PubFig, (d) FERET, (e) AR, and (f) AT&T face databases.

## 5. EXPERIMENTS AND PERFORMANCE ANALYSIS

In this section, first we analyze the effect of radius of local neighborhood and multi resolution over LDOP descriptor; then, we compare the results with existing descriptors over seven challenging face databases; and finally, we investigate the effect of distance measures. The Chi-square distance measure is used for the experiments until or otherwise specified. In order to demonstrate the improved performance of proposed LDOP descriptor for face retrieval, we compare the proposed method with state-of-the-art face descriptors such as LBP [10], LTP [25], LDP [26], SLBP [33], LVP [32], LDGP [34] and DCP [13]. Note that, all these descriptors have shown very promising performance for facial analysis under different imaging conditions such as rotation, scale, background, blur, illumination, pose, masking, etc. We have used radius $R = 1$ for LBP, LTP, LDP, SLBP and LDGP and $R = 2$ for LVP and DCP as per the source papers of these descriptors. The number of local neighbors (i.e., $N$) at a fixed radius is 8 for all the descriptors in all the experiments.

### A. Effect of Radius and Multi Resolution

The effect of radius of local neighborhood (i.e., $R$) over LDOP descriptor is observed in this experiment over each database. Fig. 6 illustrates the results in terms of the F-score for top 10 retrieved images. The number of local neighbors (i.e., $N$) at a fixed radius is 8 for LDOP in all the experiments. In Fig. 6, the values of $R$ considered for LDOP are 2, 3, 4, 5 and 6, whereas the range of value of $R$ considered for multi resolution LDOP are 23, 24, 25, 56, 46, 36, and 26. Here the range written in two digit like $\alpha\beta$ represents the multi resolution LDOP with $R = \alpha, R = \alpha + 1, ..., R = \beta$. Over LFW database, the performance of LDOP is improving with increase in the local neighborhood area with $LDOP_{26}$ as best performing combination. It can be observed that the performance of LDOP over FERET database is better if combining lower values of R such as 23, 24, and 25. This behavior of LDOP is due to the presence of huge pose variations in the FERET database which restricts to consider very wide local neighborhood. The best result over AR database is obtained for $R = 1$. Due to the non-uniform illumination variations, the result for higher values of $R$ is not improved as directional order is getting affected much. The AT&T database is also having some amount of pose variations, thus the performance is improving upto $R = 4$ over this database. It is observed that the best performance of LDOP is obtained at different settings for different databases. Still, in the rest of experiments, we use multi resolution $LDOP_{24}$.

### B. Results Comparison

The comparison results in terms of the ARP, ARR, F-score, and ANMRR against the number of retrieved images (i.e. $\gamma$) are shown in Fig. 7 over PaSC, LFW, PubFig, FERET, AR and AT&T face databases. It is clear from this result that the performance of LDOP descriptor is outstanding as compared to the other descriptors over these challenging databases. These databases are having pose, illumination, scale, unconstraint and expression variations, thus it is deduced that LDOP is more robust and discriminative as compared to LBP, LTP, LDP, SLBP, LVP, LDGP and DCP descriptors against these variations. DCP is the second best performing method in most of the cases. The PaSC database is having very complex variations like blur, illumination and pose and it is observed that LDOP is much suitable to this database. The images in LFW and PubFig databases are taken in totally uncontrolled manner. The outperformance of proposed descriptor over LFW and PubFig databases evidences the utilization of wider neighborhood for more discriminability. The FERET database is having images with huge pose differences. The better performance using LDOP over FERET database suggests that LDOP is more robust for pose variations than state-of-the-art descriptors. Over AR database, the proposed descriptor has the comparable performance to DCP. The reason of such behavior is related to the complex illumination change and masked faces that many faces in AR database have. The performance of LDOP under complex illumination will be analyzed in next sub-section over ExtendedYale database. The proposed descriptor is also good enough for the frontal faces as its performance is outstanding among all the descriptors over AT&T database. It is also observed that the improvement using multi resolution LDOP is more significant over LFW and AT&T databases, whereas if we use LDOP at lower radius then the improvement over FERET and AR databases will be also significant as suggested by the results analysis of Fig. 6. The main problem with LBP, LTP, LDP, SLBP, LDGP and DCP is due to the lack of utilization of the relationship among the dense local neighbors. Basically, these methods use only the local neighbors at a specific radius. Whereas, the proposed LDOP method uses the dense local neighborhood. Moreover, most of the existing descriptors miss to utilize the relationship among local neighbors. The LVP descriptor encodes the relationship between the differences of intensity at different directions, whereas, the proposed LDOP descriptor encodes the directional ordering pattern, whereas. Unlike the LDOP descriptor, the LVP descriptor works on high-order image space. The dimensions of the descriptors are as follows: LBP (256), LTP (512), LDP (1024), SLBP (256), LVP (1024), LDGP (65), DCP (512), and LDOP (768). However, the performance of LDOP is superior. It points out that the dimension alone is not the key factor that influences the discriminativeness of LDOP, because LDP and LVP has higher dimension than proposed method.

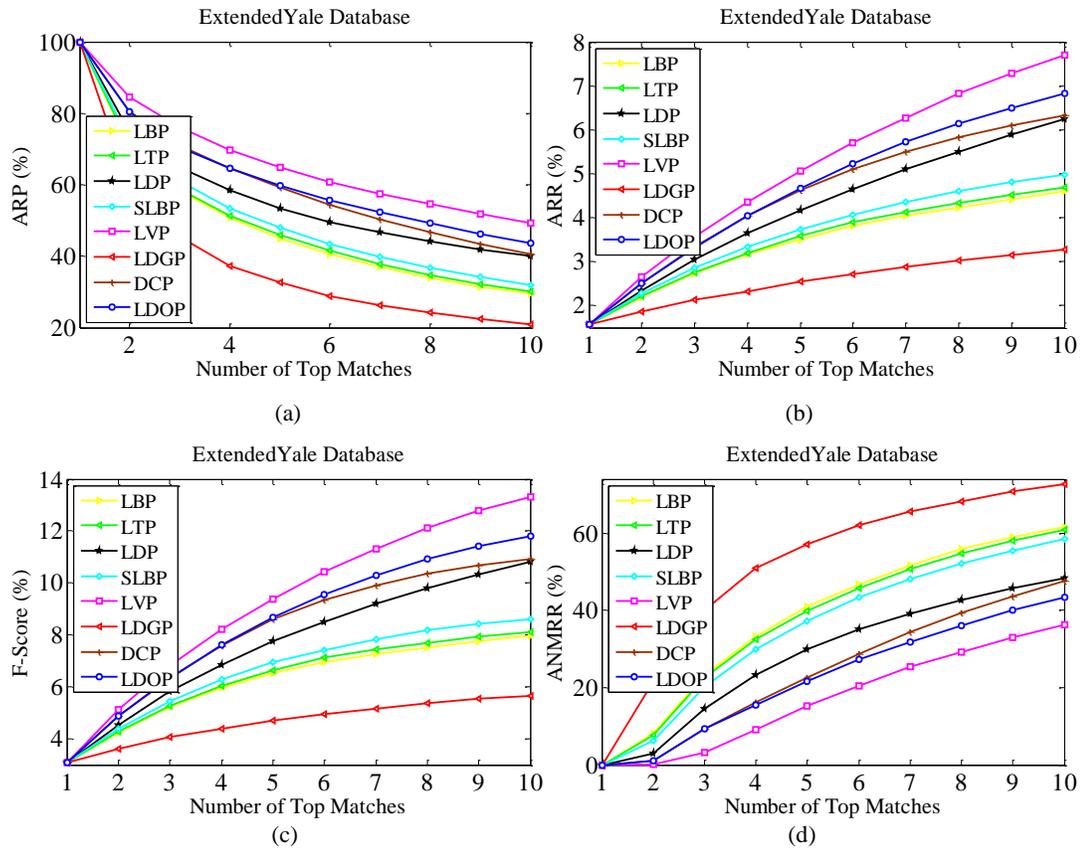

Fig.8. The performance of descriptors under extreme lighting conditions over ExtendedYale database in terms of the (a) ARP, (b) ARR, (c) F-Score, and (d) ANMRR in (%). The performance of proposed descriptor is the second best.

Table 1: The ARP (%) for $\gamma = 5$ over PaSC, LFW, PubFig, FERET, AR, AT&T, and ExtendedYale face datasets using LDOP descriptor with different distance measures such as Euclidean, Cosine, L1, D1, and Chi-square. The best result for a database is presented in bold.

| Datasets | Distance Measures | | | | |
|---|---|---|---|---|---|
| | Euclidean | Cosine | L1 | D1 | Chisq |
| PaSC | 31.34 | 32.33 | 36.49 | 36.80 | **38.16** |
| LFW | 33.31 | 33.67 | 38.19 | 38.47 | **40.15** |
| PubFig | 35.71 | 36.39 | 41.91 | 42.37 | **44.34** |
| FERET | 61.46 | 63.75 | 74.26 | 75.39 | **76.55** |
| AR | 44.49 | 45.55 | 52.02 | 52.72 | **54.55** |
| AT&T | 92.15 | 92.75 | 94.30 | **94.40** | 94.05 |
| ExtendedYale | 48.08 | 50.45 | 55.23 | 56.15 | **59.84** |

**C. Performance under Severe Illumination Changes**

Even though the intensity order is invariant to only uniform illumination change, we experiment using proposed method under non-uniform illumination change also. We conduct an experiment over ExtendedYale face database to test the robustness of proposed descriptor towards non-uniform and severe illumination changes. The results in terms of the ARP, ARR, F-Score and ANMRR are presented in Fig. 8 over ExtendedYale face database. Despite of the sensitivity of intensity order towards non-uniform intensity changes, LDOP is the second best performing descriptor over ExtendedYale face database. Moreover, its performance is significantly improved as compared to DCP for more number of retrieved images.

Pre-print: Published in Multimedia Tools and Applications, SpringerThe performance of LVP is better in this scenario due to the non-uniform light change under extreme lighting conditions. The LDOP uses the order which may not be preserved in case of non-uniform light change. This is one of the limitations of proposed method.

### D. Effect of Distance Measure

We also investigate the effect of different distances over the performance of proposed LDOP descriptor over each database in Table 1. Euclidean, Cosine, L1, D1 and Chi-square distances are used. In this table, the results are presented using ARP values in percentage for top 5 numbers of retrieved images. It is found that the ARP values using Chi-square distance measure is more as shown in bold in Table 1 over each database except AT&T. Thus, it is suggested to use the Chi-square distance with LDOP descriptor for face matching.

## 6. CONCLUSION

A local directional intensity order pattern (LDOP) based descriptor is proposed in this paper for robust face retrieval. The LDOP concept facilitates to utilize wider local neighbors without increasing the dimension of the descriptor. The directional intensity order indexes are used to encode the relationship among directional neighbors. This also provides a way to converge the wider neighborhood into the compact form. The center pixel values are transformed in the range of directional order indexes to find the LDOP pattern. The LDOP is also investigated using multi resolution mechanism. The retrieval results are compared over seven benchmark face databases (i.e., PaSC, LFW, PubFig, FERET, AR, AT&T and ExtendedYale) having different kind of variations such as pose, illumination, scale and expressions. Promising performance of proposed LDOP descriptor is observed as compared to the seven recent state-of-the-art face descriptors. The experimental results suggest that LDOP is more robust to the intra-class variations and more discriminative to the inter-class similarities. It is investigated that the performance of LDOP is better with lower radius in case of huge pose variations and non-uniform illumination changes are present in the images. It is also found that the results are better with Chi-square distance. In future, the LDOP descriptor can be used to analyze the facial expressions. It can be extended for the facial analysis by considering the color information also.

## ACKNOWLEDGMENT

This research is funded by IIIT Sri City, India through the Faculty Seed Research Grant## References

[1] G.B. Huang, M. Ramesh, T. Berg, and E. Learned-Miller E, "Labeled faces in the wild: A database for studying face recognition in unconstrained environments," *Technical Report 07-49*, University of Massachusetts Amherst, 2007.
[2] N. Kumar, A.C. Berg, P.N. Belhumeur, and S.K. Nayar, "Attribute and simile classifiers for face verification," *IEEE 12th International Conference on Computer Vision*, pp. 365-372, 2009.
[3] R. Beveridge, H. Zhang, B. Draper et al., "Report on the fg 2015 video person recognition evaluation," *IEEE International Conference on Automatic Face and Gesture Recognition*, 2015.
[4] C. Ding, C. Xu, and D. Tao, "Multi-task pose-invariant face recognition," *IEEE Transactions on Image Processing*, vol. 24, no. 3, pp. 980-993, 2015.
[5] M. Kan, S. Shan, H. Zhang, S. Lao, and X. Chen, "Multi-view discriminant analysis," *IEEE Transactions on Pattern Analysis and Machine Intelligence*, vol. 38, no. 1, pp. 188-194, 2016.
[6] W. Zhao, R. Chellappa, P.J. Phillips, and A. Rosenfeld, "Face recognition: A literature survey," *ACM Computing Surveys*, vol. 35, no. 4, pp. 399-458, 2003.
[7] X. Zhang and Y. Gao, "Face recognition across pose: A review," *Pattern Recognition*, vol. 42, no. 11, pp. 2876-2896, 2009.
[8] C. Ding and D. Tao, "A comprehensive survey on pose-invariant face recognition," *ACM Transactions on Intelligent Systems and Technology*, vol. 7, no. 3, pp. 37, 2016.
[9] J. Wright, A. Y. Yang, A. Ganesh, S. S. Sastry, and Y. Ma, "Robust face recognition via sparse representation," *IEEE Transactions on Pattern Analysis and Machine Intelligence*, vol. 31, no. 2, pp. 210–227, Feb. 2009.
[10] T. Ahonen, A. Hadid, and M. Pietikainen, "Face description with local binary patterns: Application to face recognition," *IEEE Transactions on Pattern Analysis and Machine Intelligence*, vol. 28, no. 12, pp. 2037-2041, 2006.This paper is published in Multimedia Tools and Applications, Springer. Cite this article as: S.R. Dubey and S. Mukherjee, "LDOP: Local Directional Order Pattern for Robust Face Retrieval", Multimedia Tools and Applications, 2019. https://doi.org/10.1007/s11042-019-08370-x13